\documentclass[conference]{IEEEtran}
\usepackage[english]{babel}
\usepackage{mathtools}
\usepackage{amssymb, amsthm}
\usepackage{hyperref}
\usepackage{cite}
\usepackage{todonotes}
\usepackage{tikz}
\usepackage{caption}
\usepackage{graphicx}
\usepackage{epstopdf}
\usepackage{color}
\usepackage{wrapfig}
\usepackage{caption}
\usepackage{pgfplots}

\pgfplotsset{every x tick label/.append style={font=\tiny, yshift=0.5ex}}
\pgfplotsset{every y tick label/.append style={font=\tiny, xshift=0.5ex}}
\pgfplotsset{every axis legend/.append style={font={\tiny, xshift=0.5ex, yshift=0.5ex}}}

\ifCLASSOPTIONcompsoc
    \usepackage[caption=false,font=normalsize,labelfont=sf,textfont=sf]{subfig}
\else
    \usepackage[caption=false,font=footnotesize]{subfig}
\fi

\usetikzlibrary{arrows,positioning,automata,bayesnet}

\title{Compressive Sensing Approaches for Autonomous Object Detection in Video Sequences}

\author{\IEEEauthorblockN{Danil Kuzin}
\IEEEauthorblockA{The University of Sheffield\\
Sheffield, UK\\
Email: dkuzin1@sheffield.ac.uk}
\and
\IEEEauthorblockN{Olga Isupova}
\IEEEauthorblockA{The University of Sheffield\\
Sheffield, UK\\
Email: o.isupova@sheffield.ac.uk}
\and
\IEEEauthorblockN{Lyudmila Mihaylova}
\IEEEauthorblockA{The University of Sheffield\\
Sheffield, UK\\
Email: l.s.mihaylova@sheffield.ac.uk}
}

\begin{document}

\maketitle

\begin{abstract}
Video analytics requires operating with large amounts of data. Compressive sensing allows to reduce the number of measurements required to represent the video using the prior knowledge of sparsity of the original signal, but it imposes certain conditions on the design matrix. The Bayesian compressive sensing approach relaxes the limitations of the conventional approach using the probabilistic reasoning and allows to include different prior knowledge about the signal structure. This paper presents two Bayesian compressive sensing methods for autonomous object detection in a video sequence from a static camera. Their performance is compared on the real datasets with the non-Bayesian greedy algorithm. It is shown that the Bayesian methods can provide the same accuracy as the greedy algorithm but much faster; or if the computational time is not critical they can provide more accurate results.
\end{abstract}

\section{Introduction}
The significant developments in the field of sparse models during the last decades lead to the opening of the new research and application fields. 

One of the first application for sparse modelling is the linear regression problem where $l_0$ and $l_1$-norm regularisation is considered. The latter has the advantage that a regulariser term is convex, while it has not so obvious sparse interpretation~\cite{Bach2012}.   

Sparse modelling is further developed in the field of signal processing in compressive sensing~\cite{Candes2008intro}, where the main idea is to minimise the number of measurements of the signal without loss of the decoding accuracy. Compressive sensing concerns the two main problems: selecting the optimal design matrix and solving ill-posed regression, that arises in the original signal decoding from the measurements~\cite{Mihaylova2014}. 

The idea of sparse Bayesian modelling is mentioned in~\cite{Tibshirani96}. This imposes the sparsity-inducing Laplace prior on the data, but does not give the inference for the whole distribution, only a maximum aposteriori probability estimate. The full inference to this model is provided in~\cite{Seeger08}, using the Expectation Propagation~(EP) technique. Another work is~\cite{Tipping2001}, where the prior is modified to the hierarchical Gauss-Gamma distribution. These models are used as a basis for Bayesian compressive sensing in~\cite{Carin2008bcs} and~\cite{Seeger08cs}. 

The recent monograph~\cite{Bach14} presents the sparse modelling application for image and video processing. One of the essential problems in video processing is foreground detection which is mostly solved by background subtraction. Background subtraction aims to distinguish foreground~(moving objects) from background~(static ones). Sparseness is natural for the background subtraction problem as the foreground objects occupy the small regions on a frame. Background subtraction hence represents a natural application area for sparse modelling. 

The idea to apply compressive sensing for background subtraction is originally proposed in~\cite{Cevher2008} and developed in~\cite{Warnell2014}. In contrast to these works in this paper we focus on the sparse Bayesian methods for background subtraction and the comprehensive comparison of these methods with the conventional compressive sensing one. 

The contribution of this paper is in applying the Bayesian compressive sensing approach for the background subtraction problem. As far as the authors know, this approach for moving object detection has not been considered yet. Also several algorithms are overviewed and compared to evaluate their applicability in different situations.

This paper is organised as follows. In Section~\ref{sec:framework} the proposed model is explained. The experimental results are represented in Section~\ref{sec:experiments}. Section~\ref{sec:conclusion} concludes the paper and discusses the future work.  
 
\section{Framework}
\label{sec:framework}
Assume that we have a static camera and we can acquire a frame $\mathbf{B}\in\mathbb{R}^{n_1 \times n_2}$ from the camera that is referenced as the background. The video from the camera consists of the sequential frames $\mathbf{V}_k \in\mathbb{R}^{n_1 \times n_2},\,k\in \{1,\ldots,K\}$. The aim is to estimate the mask of the foreground objects in these frames. 

\subsection{Video preprocessing}
We convert the source video frames to greyscale. The background frame $\mathbf{B}$ is converted to a vector $\mathbf{b} \in \mathbb{R}^n$, the video frames $\mathbf{V}_k$ are converted to vectors $\mathbf{v}_k \in \mathbb{R}^n$, where $n=n_1 n_2$.

\subsection{Compressive sensing}
Typically the foreground objects take only a part of the image. Therefore the foreground mask $\mathbf{f}_k=\mathbf{v}_k-\mathbf{b}$ has many values that are close to zero. This intuition can be represented as an assumption of
\begin{equation}
\|\mathbf{f}_k\|_{l_0}\le s \ll n,
\end{equation}
$l_0$-pseudonorm is the number of non-zero elements of a vector.
 
We apply the compressive sensing theory to this problem. It reduces the number of measurements that need to be taken~\cite{Candes2008intro} and also the results may be denoised~\cite{Bach14}. The values of the foreground mask are estimated based on the set of the compressed measurements $\mathbf{g}_k \in \mathbb{R}^{s}$: 
\begin{equation}
\mathbf{g}_k = \mathbf{\Phi} \mathbf{f}_k, 
\end{equation}
where the design matrix\,$\mathbf{\Phi}\in\mathbb{R}^{s \times n}$ consists of i.i.d Gaussian variables. It is selected according to the method proposed in~\cite{Baraniuk08}.

Since $\mathbf{f}_k=\mathbf{v}_k-\mathbf{b}$, the estimates of the coefficients $\mathbf{g}_k$ can be done on the acquisition step as 
\begin{equation}
\label{eq:linear_system}
\mathbf{g}_k = \mathbf{\Phi} \mathbf{f}_k = \mathbf{\Phi} \mathbf{v}_k - \mathbf{\Phi} \mathbf{b}
\end{equation}
The vectors $\mathbf{\Phi} \mathbf{b}$ and $\mathbf{\Phi} \mathbf{v}_k$ are the linear combinations of the pixels of the video frames, therefore a single pixel camera may be used. The problem of the foreground mask reconstruction is more difficult. 

The linear system (\ref{eq:linear_system}) is underdetermined when $n > s$ therefore an infinite amount of solutions exists. The problem can be determined by the regulariser imposing in the assumption that the signal $\mathbf{f}_k$ has a sparse structure. The common regularisers that are used in compressive sensing are minimisers of the~$l_p$~norm, where $p < 2$.

The conventional methods to solve such systems are following \cite[Chapter 13]{Murphy2012}: 
\begin{itemize}
\item \textit{$l_0$ - minimisation.} The greedy algorithms based on least squares estimates, stochastic search, variational inference;
\item \textit{$l_1$ - minimisation.} Coordinate descent, LARS, the proximal and gradient projection methods;
\item \textit{Non-convex minimisation.} Bridge regression, hierarchical adaptive lasso
\end{itemize} 

In this paper we will focus on the Bayesian methods~\cite{Carin2008bcs,Carin2009mcs} and compare them with orthogonal matching pursuit~(OMP)~\cite{Mallat93}, that is a greedy algorithm for $l_0$-minimisation. The following represents the brief review of these methods.

\subsubsection{Bayesian compressive sensing (BCS)}

\begin{figure}[!t]
\begin{center}
\includegraphics{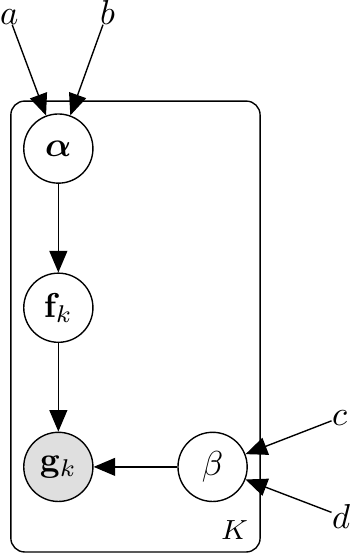}
\caption{The graphical model for Bayesian compressive sensing}
\label{pic:bcs_gm}
\end{center}
\end{figure}

\begin{figure}[!t]
\begin{center}
\includegraphics{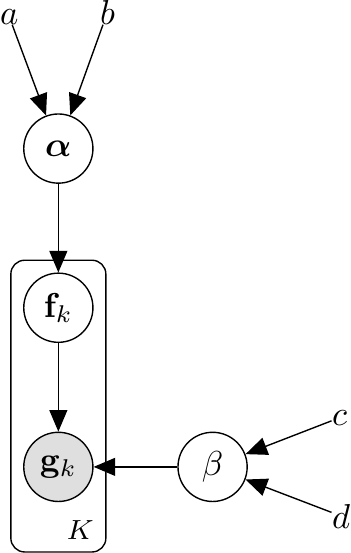}
\caption{The graphical model for multitask Bayesian compressive sensing}
\label{pic:mt_gm}
\end{center}
\end{figure}

The system (\ref{eq:linear_system}) is reformulated as a linear regression model in \cite{Carin2008bcs}:
\begin{equation}
\mathbf{g}_k = \mathbf{\Phi}\mathbf{f}_k + \boldsymbol{\xi},
\end{equation}
where $\boldsymbol{\xi}$ is a vector which elements are the independent noise from the Gaussian distribution: $\xi_i\sim\mathcal{N}(\xi_i;0,\beta^{-1})$. Therefore the likelihood can be expressed as
\begin{equation}
p(\mathbf{g}_{k} | \mathbf{f}_k, \beta) = \prod\limits_{i=1}^n\mathcal{N}(g_{i, k};\mathbf{\Phi}_i\mathbf{f}_k,\beta^{-1}),
\end{equation}
where $g_{i, k}$ is the $i$-th element of the vector~$\mathbf{g}_k$, $\mathbf{\Phi}_i$ -- the $i$-th row of the matrix~$\mathbf{\Phi}$.

To implement the full Bayesian approach, the prior distributions are imposed on all parameters:
\begin{equation}
p(\mathbf{f}_k | \boldsymbol\alpha) = \prod\limits_{i=1}^n\mathcal{N}(f_{i, k};0,\alpha_i^{-1}),
\end{equation}
where $f_{i, k}$ is the $i$-th element of the vector $\mathbf{f}_k$, $\boldsymbol{\alpha}$ is a prior parameter vector, $\alpha_i$ is the $i$-th element of the vector $\boldsymbol{\alpha}$;
\begin{equation}
p(\boldsymbol\alpha) = \prod\limits_{i=1}^n\Gamma(\alpha_i;a,b),
\end{equation}
\begin{equation}
p(\beta) = \Gamma(\beta;c,d),
\end{equation}
where $\Gamma(\cdot)$ denotes the Gamma distribution. The values of the hyperparameters $a, b, c, d$ are set uniform and close to zero.

According to the Bayes rule the posterior distribution can be written as follows:
\begin{equation}
p(\mathbf{f}_k, \boldsymbol\alpha, \beta | \mathbf{g}_k) = \frac{p(\mathbf{g}_k|\mathbf{f}_k, \boldsymbol\alpha, \beta)p(\mathbf{f}_k, \boldsymbol\alpha, \beta)}{p(\mathbf{g}_k)},
\end{equation}
where $p(\mathbf{g}_k|\mathbf{f}_k, \boldsymbol\alpha, \beta)$ is the likelihood term, $p(\mathbf{f}_k, \boldsymbol\alpha, \beta)$ is the prior term, $p(\mathbf{g}_k)$ is the evidence term. The latter can be expressed as: 
\begin{equation}
p(\mathbf{g}_k) = \int\limits_{\mathbf{f}_k, \boldsymbol\alpha, \beta} p(\mathbf{g}_k|\mathbf{f}_k, \boldsymbol\alpha, \beta) \, p(\mathbf{f}_k, \boldsymbol\alpha, \beta) \, d\mathbf{f}_k \, d\boldsymbol\alpha \, d\beta
\end{equation}
This integral is intractable, therefore some kind of approximation should be used.

In Bayesian compressive sensing \cite{Carin2008bcs} the decomposition of the posterior probability into the product of the tractable and intractable probabilities is used and the intractable one is approximated with the delta-function in its mode:
\begin{equation}
\label{CsPosterior}
p(\mathbf{f}_k, \boldsymbol\alpha, \beta | \mathbf{g}_k) = p(\mathbf{f}_k |\mathbf{g}_k, \boldsymbol\alpha, \beta)p(\boldsymbol\alpha, \beta | \mathbf{g}_k)
\end{equation}

The Bayes rule for the first term of (\ref{CsPosterior}) is as follows: 
\begin{equation}
p(\mathbf{f}_k | \mathbf{g}_k, \boldsymbol\alpha, \beta) =  \frac{p(\mathbf{g}_k | \mathbf{f}_k, \beta) p(\mathbf{f}_k | \boldsymbol\alpha)}{p(\mathbf{g}_k | \boldsymbol\alpha, \beta)}
\end{equation}
These are all the Gaussians, so the probability $p(\mathbf{f}_k | \boldsymbol\alpha, \beta, \mathbf{g}_k)$ can be calculated straightforwardly. It is the Gaussian distribution with the parameters 
\begin{align}
\label{CsLikelihoodParametersSigma}
\mathbf{\Sigma} &= (\beta\mathbf{\Phi}^{\top}\mathbf{\Phi} + \mathbf{A})^{-1}, \\
\label{CsLikelihoodParametersMu}
\boldsymbol\mu &= \beta\mathbf{\Sigma}\mathbf{\Phi}^{\top}\mathbf{g}_k,
\end{align}
where $\mathbf{A} = \text{diag}(\alpha_1, \ldots, \alpha_n)$.

The second term of the posterior probability (\ref{CsPosterior}) can be expressed as:
\begin{equation}
p(\boldsymbol\alpha, \beta | \mathbf{g}_k) = \frac{p(\mathbf{g}_k | \boldsymbol\alpha, \beta) p(\boldsymbol\alpha) p(\beta)}{p(\mathbf{g}_k)}
\end{equation}

As it has been already shown, the denominator here is not tractable. The most probable values of $\boldsymbol\alpha, \beta$ are used. The hyperpriors are uniform, therefore only the term $p(\mathbf{g}_k | \boldsymbol\alpha, \beta)$ needs to be maximised:
\begin{equation}
\label{eq:posterior_alphabeta}
p(\mathbf{g}_k | \boldsymbol\alpha, \beta) = \int p(\mathbf{g}_k |\mathbf{f}_k, \beta) p(\mathbf{f}_k | \boldsymbol\alpha) d\mathbf{f}_k
\end{equation}
Maximisation of (\ref{eq:posterior_alphabeta}) w.r.t $\boldsymbol\alpha, \beta$ gives the following iterative process:
\begin{align}
\alpha_{i}^{new} &= \frac{\gamma_i}{\mu_i^2}, \\
(\beta^{-1})^{new} &= \frac{\|\mathbf{g}_k - \mathbf{\Phi}\mu\|^2_{l_2}}{s-\Sigma_{ii}\gamma_i},
\end{align}
where $\gamma_i = 1 - \alpha_i\Sigma_{ii}$

This process together with (\ref{CsLikelihoodParametersSigma}) - (\ref{CsLikelihoodParametersMu}) converges to the optimal estimates.

Note that 
\begin{equation}
p(f_{i, k}) = \frac{b^a\Gamma\left(a+\frac{1}{2}\right)}{(2\pi)^{\frac{1}{2}}\Gamma(a)}\left(b + \frac{f_{i, k}^{2}}{2}\right)^{-\left(a+\frac{1}{2}\right)}
\end{equation} 
This is the Student-t distribution, that has the most probable area concentrated around zero. Thereby it leads to the sparse vector $\mathbf{f}_k$.

The graphical model is displayed in Figure \ref{pic:bcs_gm}.

\subsubsection{Multitask Bayesian compressive sensing (Multitask BCS)}
In \cite{Carin2009mcs} the Bayesian method to process several signals that have a similar sparse structure is proposed. The multitask setting reduces the number of measurements that should be taken comparing to processing all the signals independently. The hyperparameter $\boldsymbol\alpha$ is considered to be shared by all the tasks. The graphical model is displayed in Figure \ref{pic:mt_gm}. 

\subsubsection{Matching Pursuit}
The greedy algorithms are proposed for the $l_0$ minimisation in \cite{Mallat93}. These methods start with a null vector and iteratively add variables to it until a convergence to a threshold.

\section{Experiments}
\label{sec:experiments}
\begin{figure*}[!t]
\centering
\subfloat[]{\includegraphics[width=\textwidth/5]{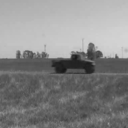}}
~
\subfloat[]{\includegraphics[width=\textwidth/5]{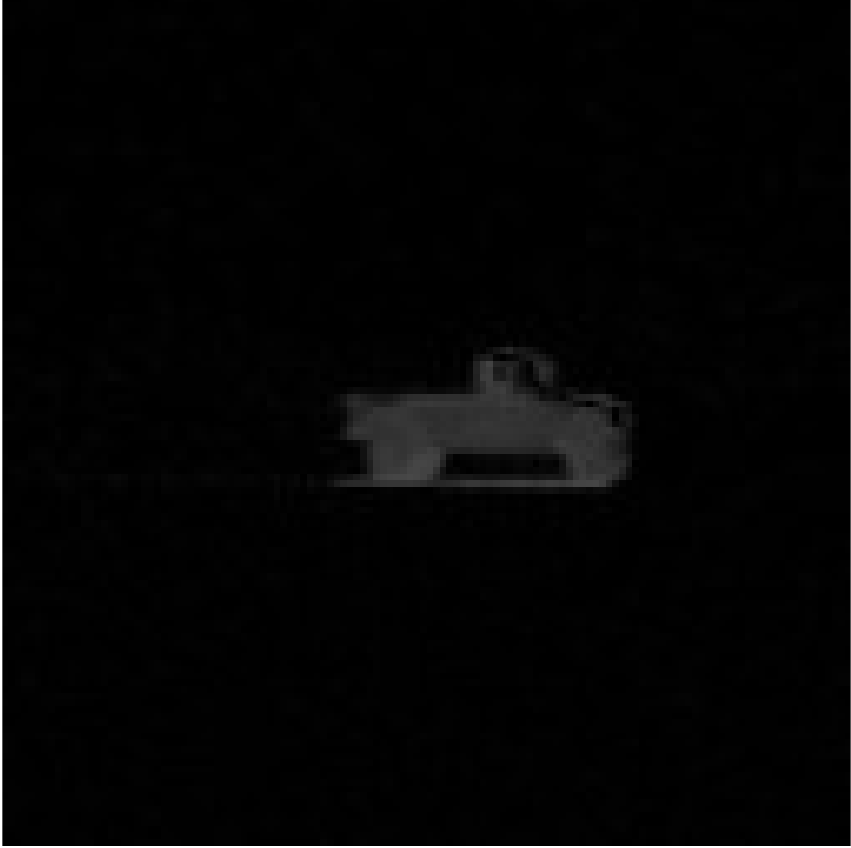}}
~
\subfloat[]{\includegraphics[width=\textwidth/5]{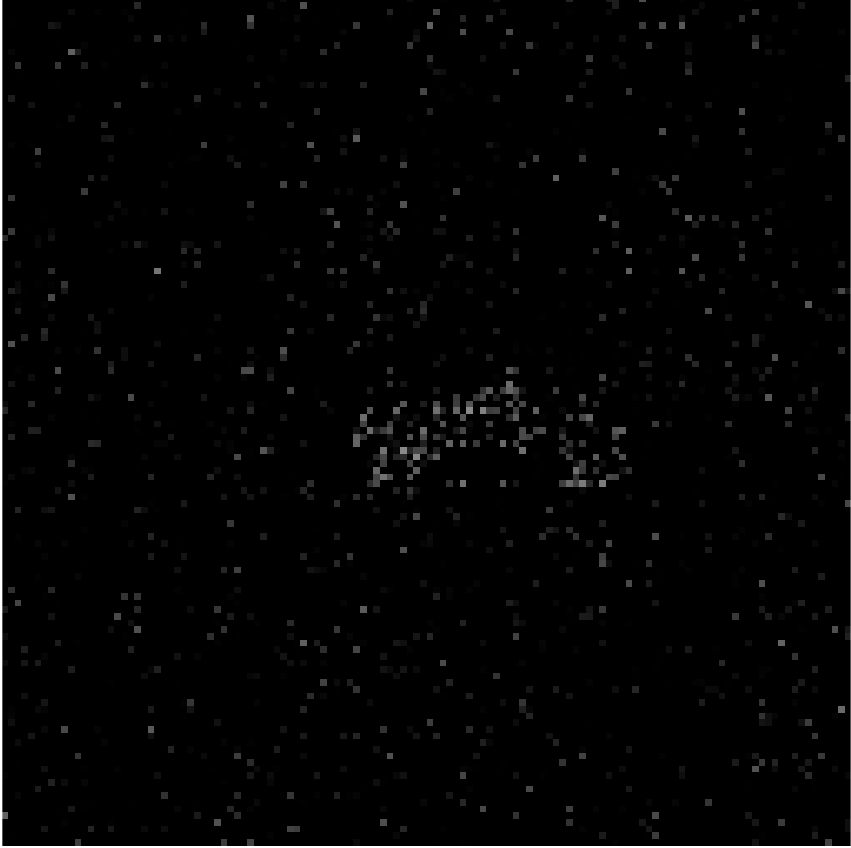}}
~
\subfloat[]{\includegraphics[width=\textwidth/5]{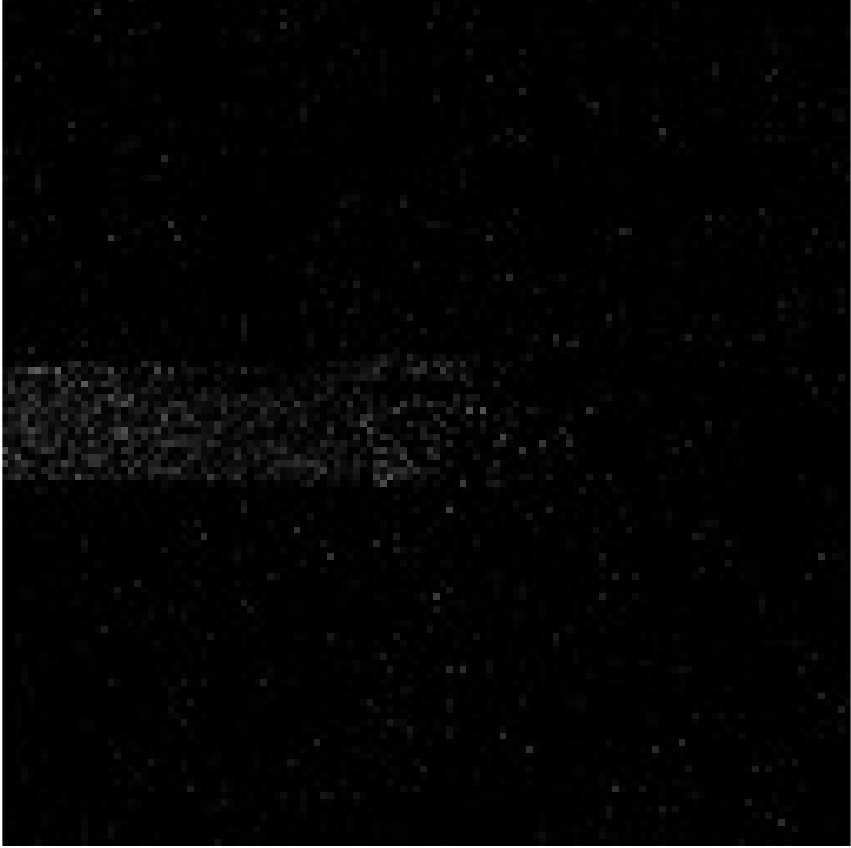}}
~
\subfloat[]{\includegraphics[width=\textwidth/5]{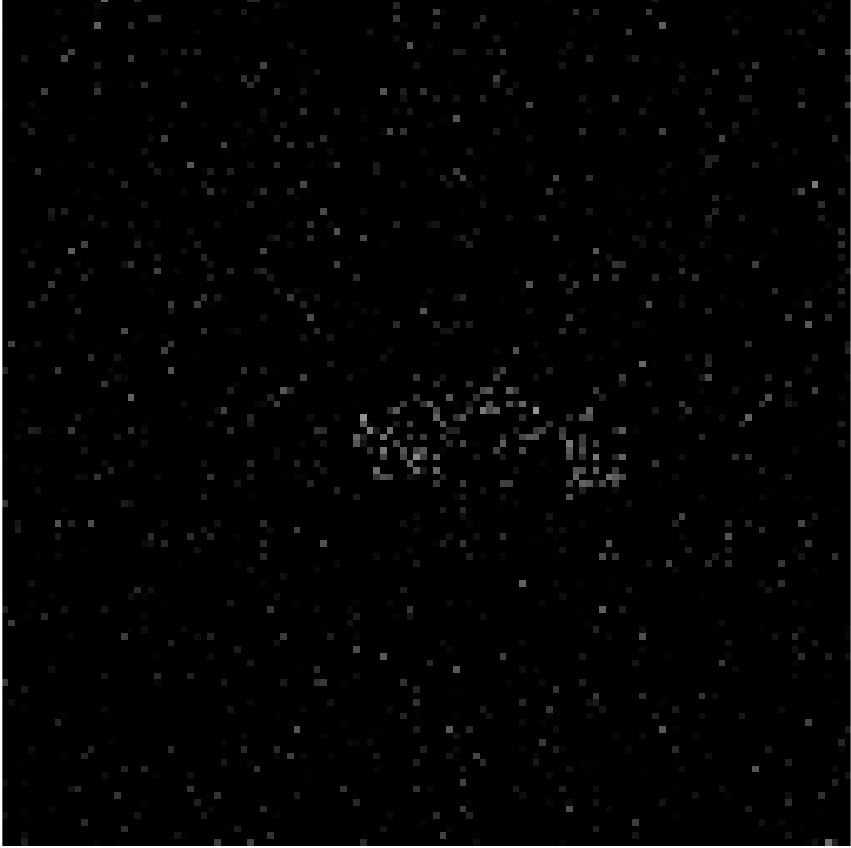}}

\subfloat[]{\includegraphics[width=\textwidth/5]{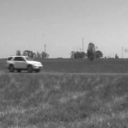}}
~
\subfloat[]{\includegraphics[width=\textwidth/5]{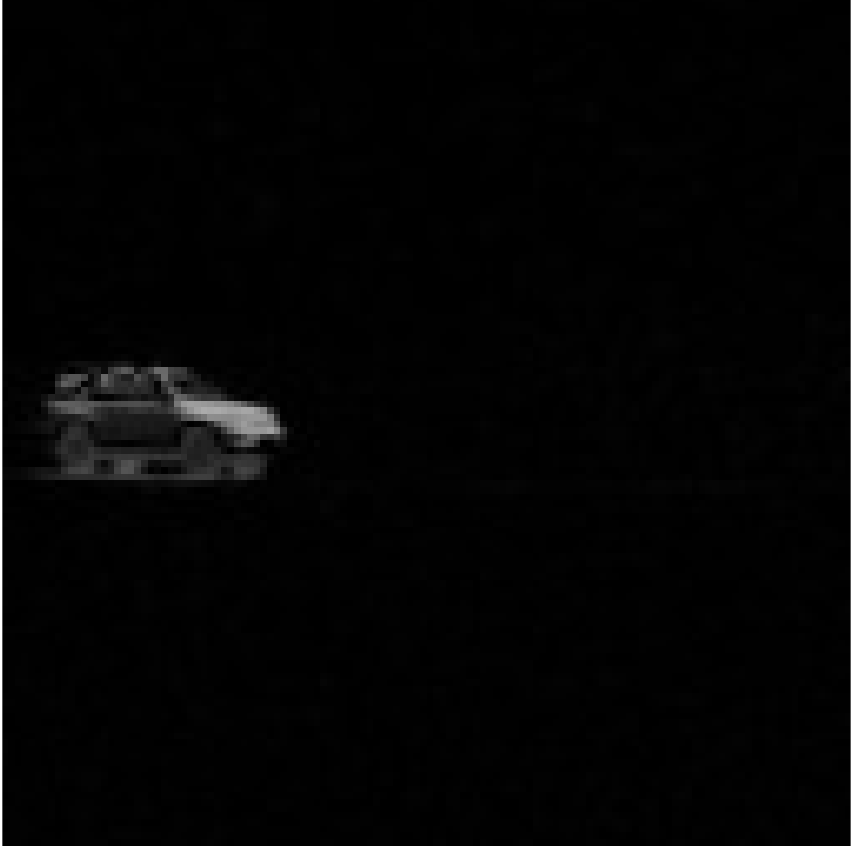}}
~
\subfloat[]{\includegraphics[width=\textwidth/5]{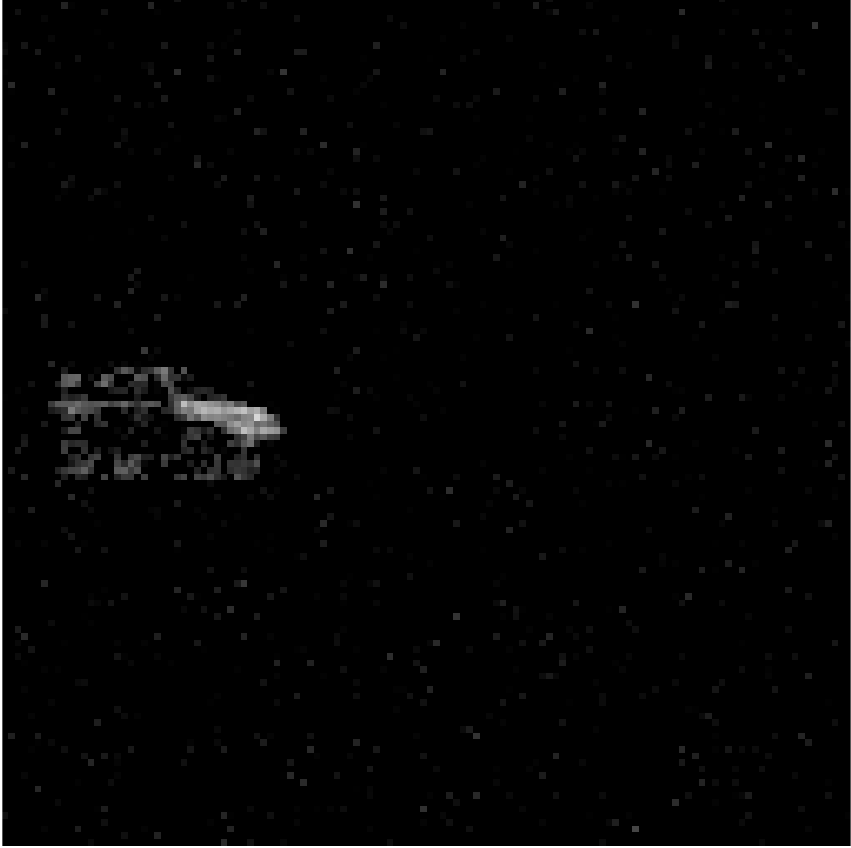}}
~
\subfloat[]{\includegraphics[width=\textwidth/5]{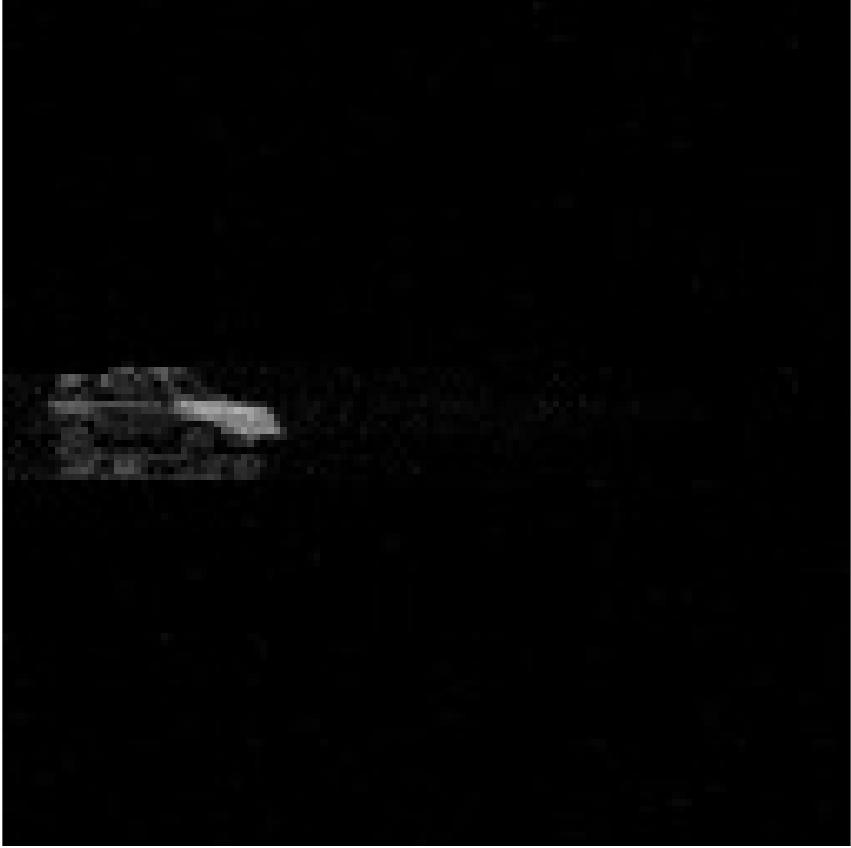}}
~
\subfloat[]{\includegraphics[width=\textwidth/5]{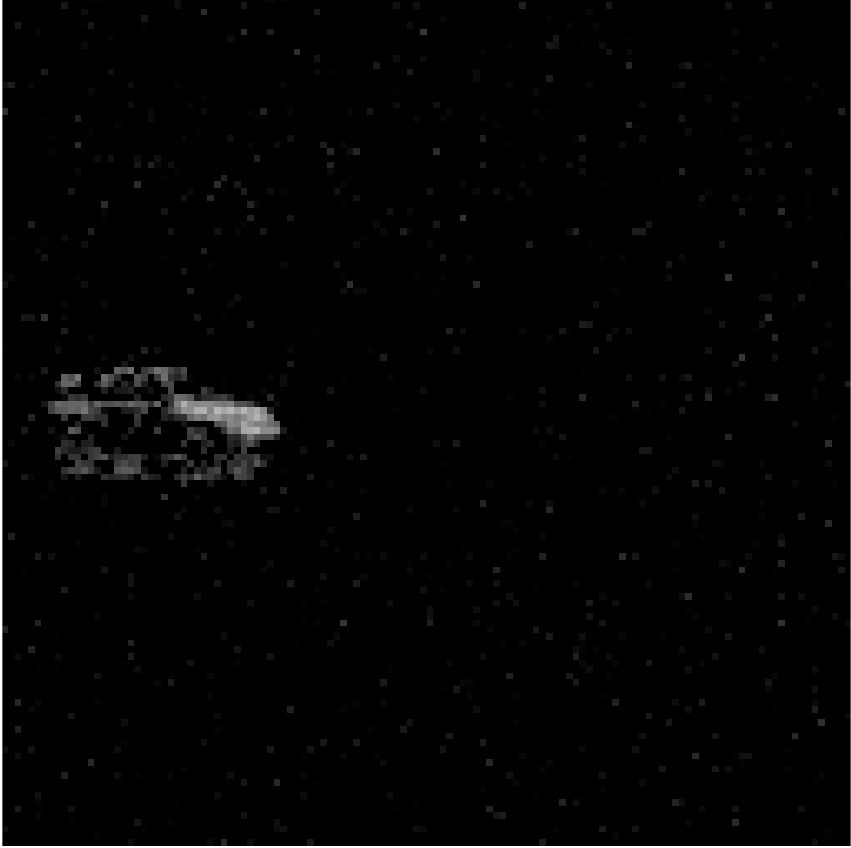}}

\subfloat[]{\includegraphics[width=\textwidth/5]{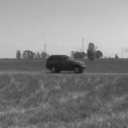}}
~
\subfloat[]{\includegraphics[width=\textwidth/5]{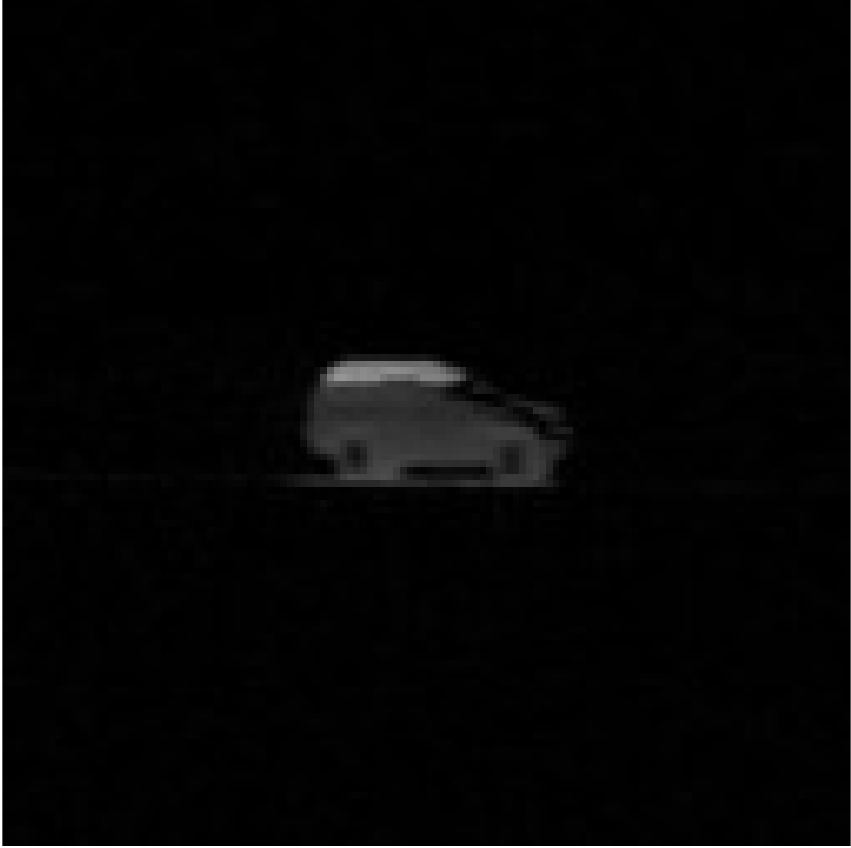}}
~
\subfloat[]{\includegraphics[width=\textwidth/5]{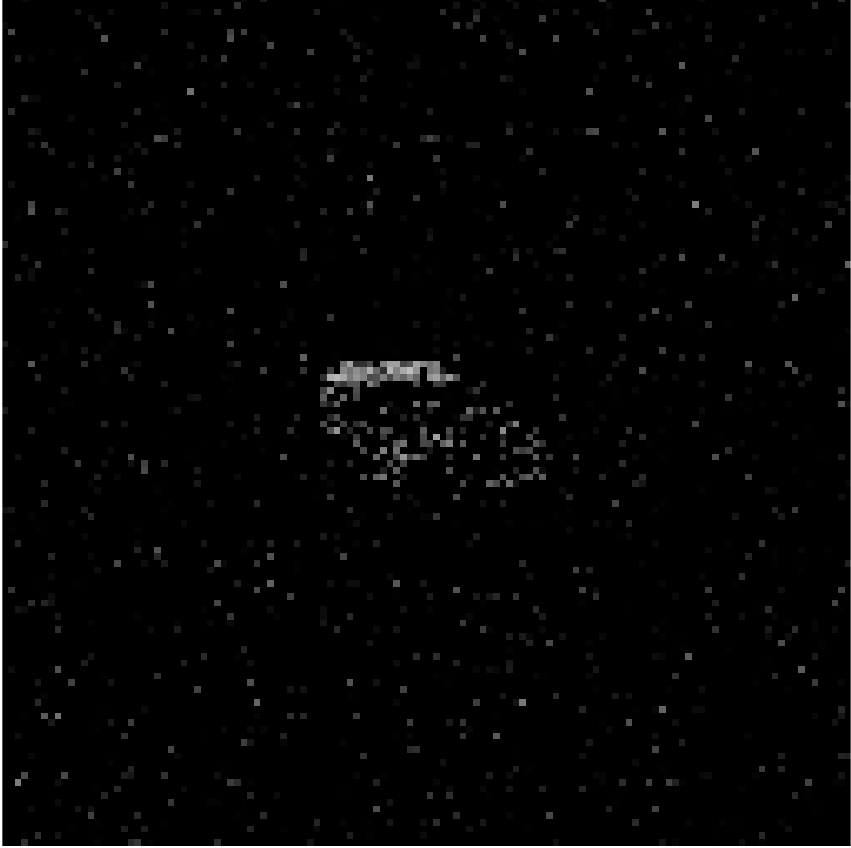}}
~
\subfloat[]{\includegraphics[width=\textwidth/5]{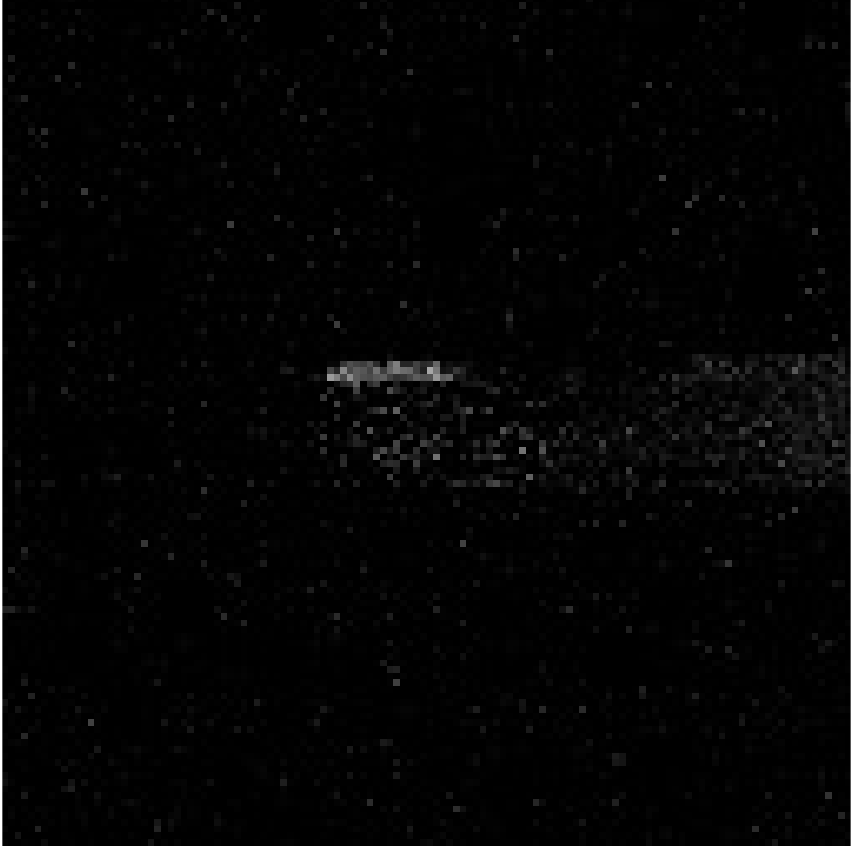}}
~
\subfloat[]{\includegraphics[width=\textwidth/5]{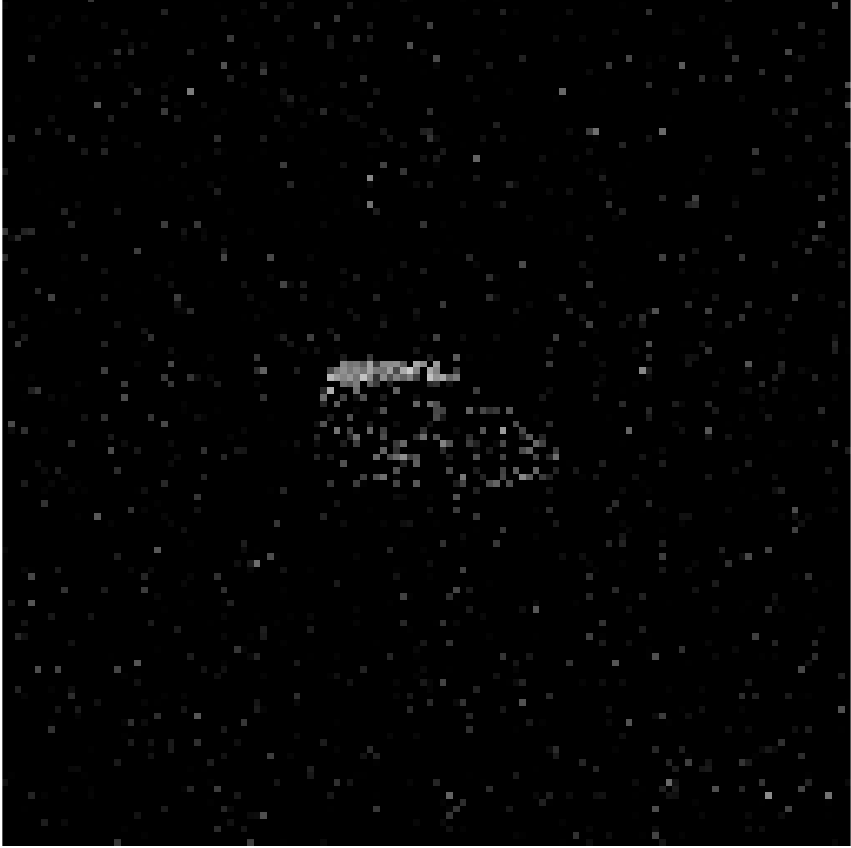}}
\caption{Comparison of the foreground restoration based on 2000 measurements by the algorithms. The three rows correspond to the three sample frames. From left to right columns: the input uncompressed frame, uncompressed background subtraction, compressed background subtraction with Bayesian compressive sensing, compressed background subtraction with multi-task Bayesian compressive sensing, compressed background subtraction with orthogonal matching pursuit}
\label{fig:results2000}
\end{figure*}

We use the Convoy dataset \cite{Warnell2014}, which consists of 260 greyscale frames and the background frame. The frames are scaled to the less resolution of $128 \times 128$ to avoid memory problems. For the multitask algorithm the batches of 40 frames are run together, while for the Bayesian compressive sensing and OMP algorithms all the frames are processed independently. There are two sets of the experiments: one with $s = 2000$ measurements and the other with $s = 5000$ measurements. For both sets of the experiments all three methods are run for 10 times with 10 different design matrices $\mathbf{\Phi}$ shared among the methods. For the quantitative comparison the median values of quality measures among these runs are presented.

\begin{figure*}[!t]
\centering
\subfloat[]{\includegraphics[width=\textwidth/5]{convoy2_040}}
~
\subfloat[]{\includegraphics[width=\textwidth/5]{pure_bs_40}}
~
\subfloat[]{\includegraphics[width=\textwidth/5]{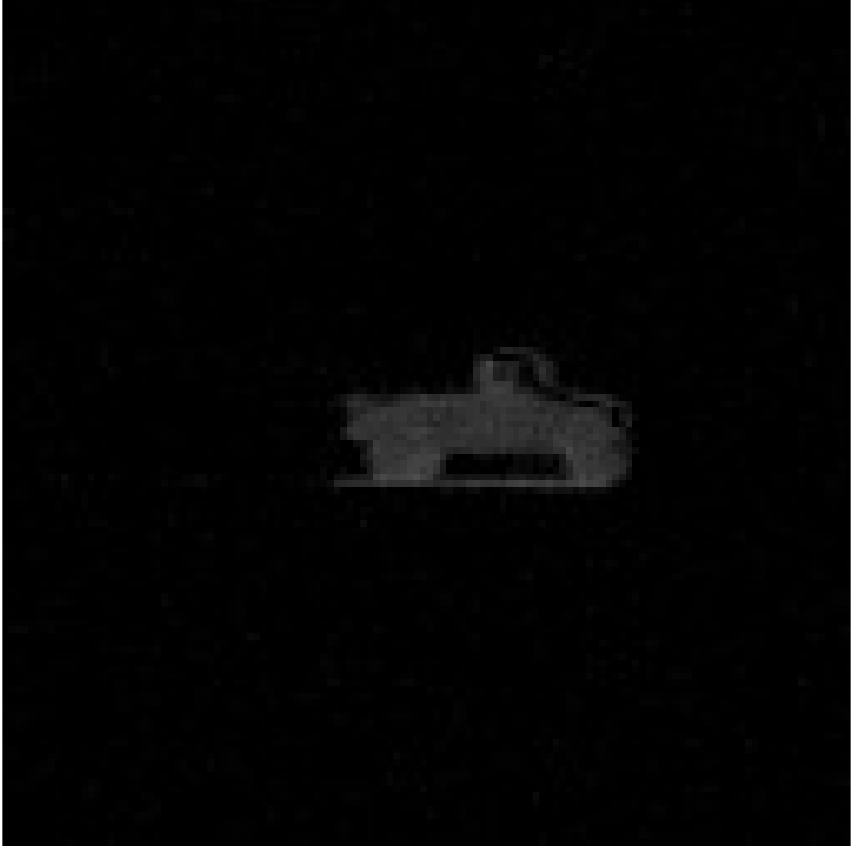}}
~
\subfloat[]{\includegraphics[width=\textwidth/5]{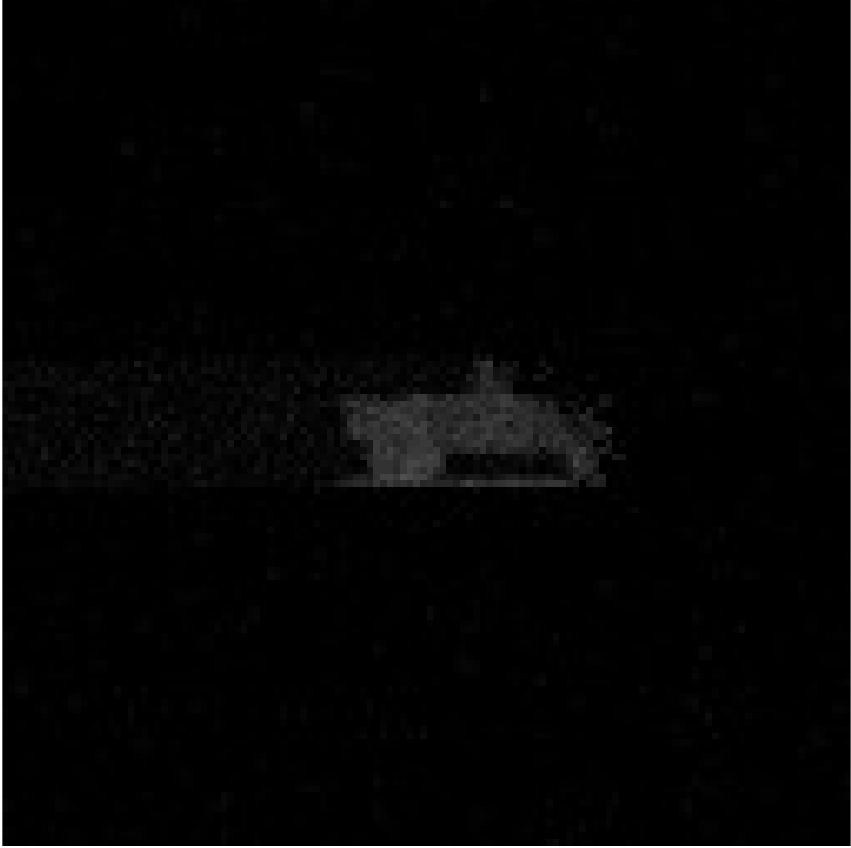}}
~
\subfloat[]{\includegraphics[width=\textwidth/5]{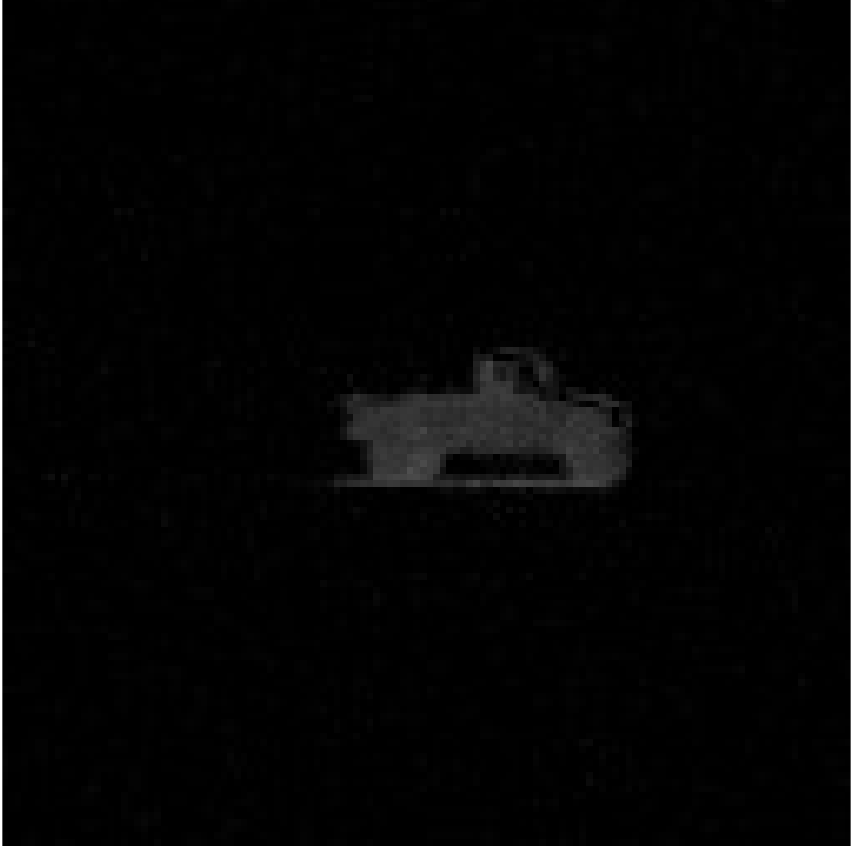}}

\subfloat[]{\includegraphics[width=\textwidth/5]{convoy2_099}}
~
\subfloat[]{\includegraphics[width=\textwidth/5]{pure_bs_99}}
~
\subfloat[]{\includegraphics[width=\textwidth/5]{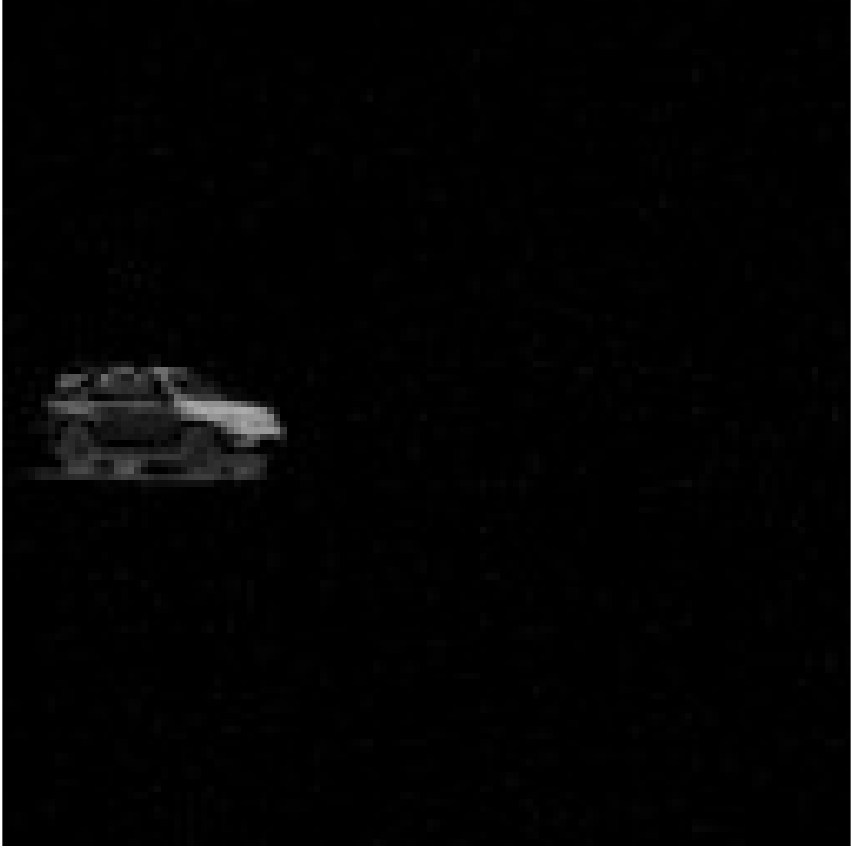}}
~
\subfloat[]{\includegraphics[width=\textwidth/5]{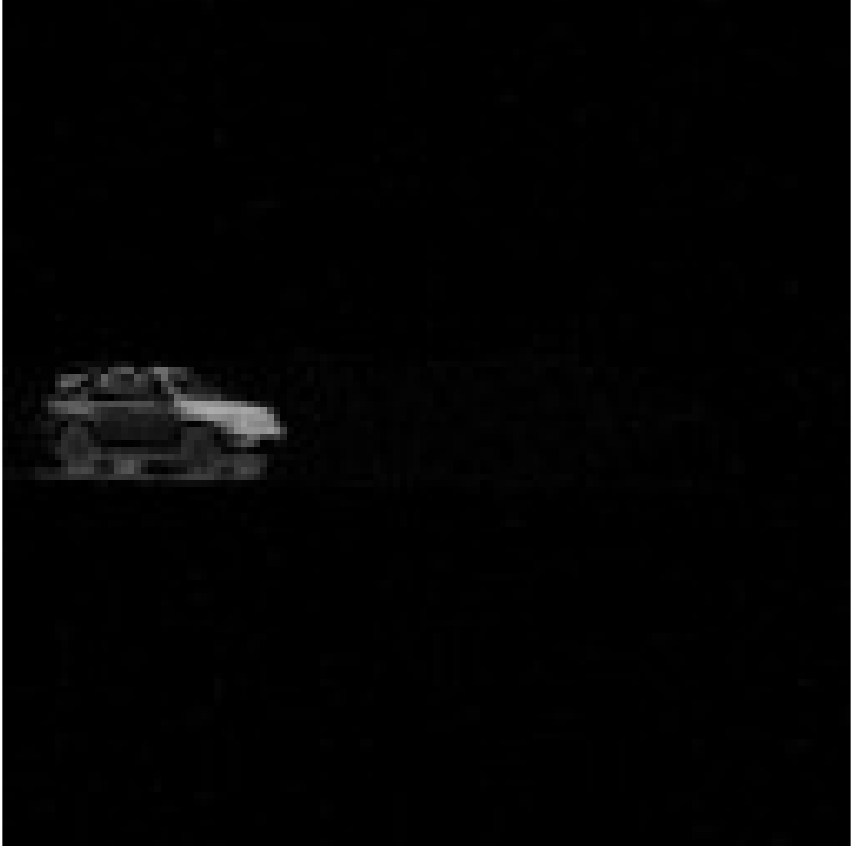}}
~
\subfloat[]{\includegraphics[width=\textwidth/5]{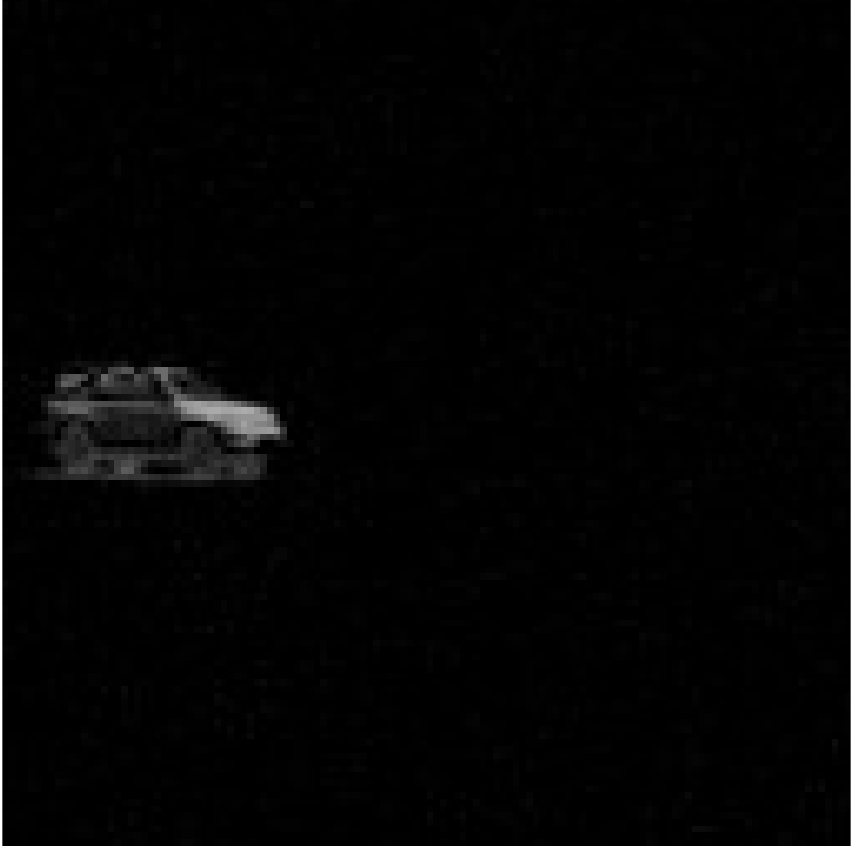}}

\subfloat[]{\includegraphics[width=\textwidth/5]{convoy2_203}}
~
\subfloat[]{\includegraphics[width=\textwidth/5]{pure_bs_203}}
~
\subfloat[]{\includegraphics[width=\textwidth/5]{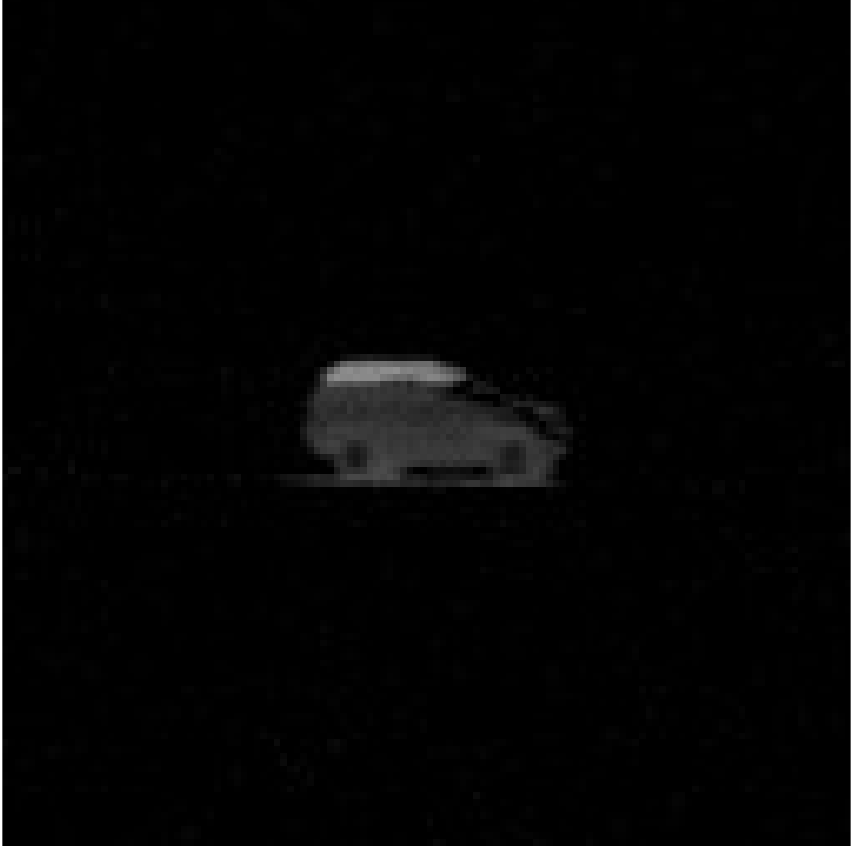}}
~
\subfloat[]{\includegraphics[width=\textwidth/5]{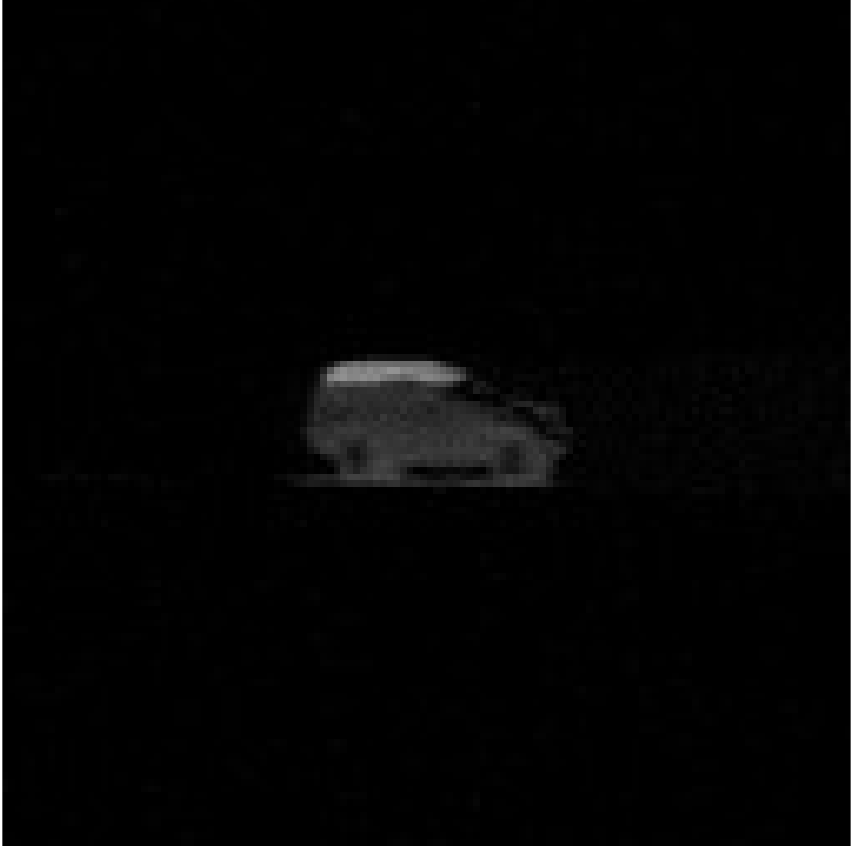}}
~
\subfloat[]{\includegraphics[width=\textwidth/5]{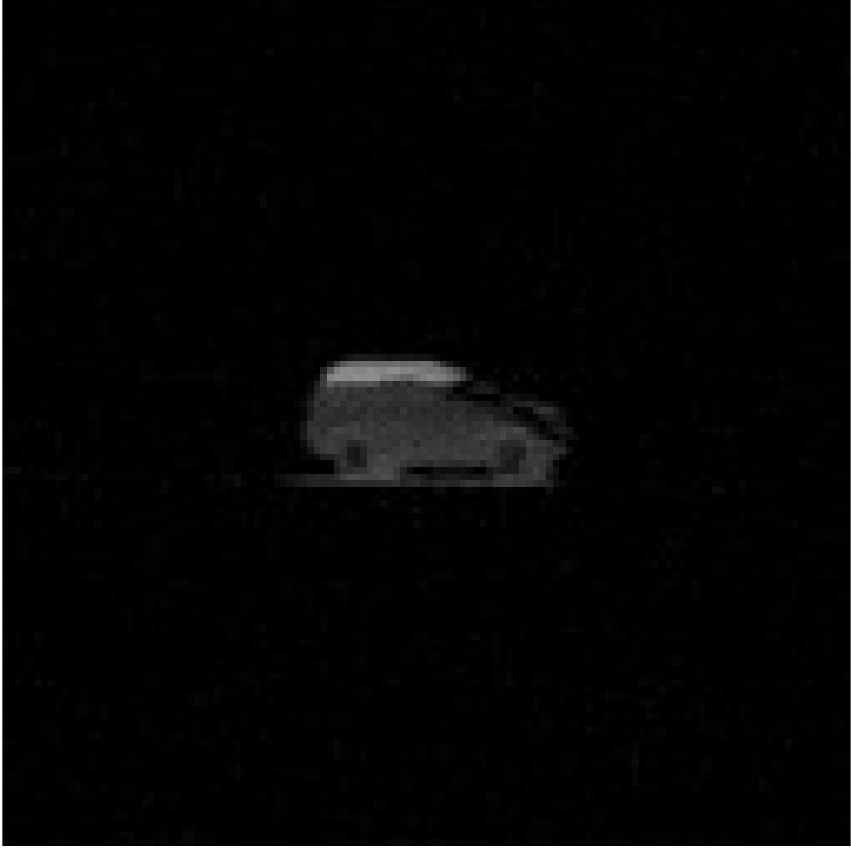}}
\caption{Comparison of the foreground restoration based on 5000 measurements by the algorithms. The three rows correspond to the three sample frames. From left to right columns: the input uncompressed frame, uncompressed background subtraction, compressed background subtraction with Bayesian compressive sensing, compressed background subtraction with multi-task Bayesian compressive sensing, compressed background subtraction with orthogonal matching pursuit}
\label{fig:results5000}
\end{figure*}

The qualitative comparison of the models with the same design matrix $\mathbf{\Phi}$ is displayed in Figures~\ref{fig:results2000}~-~\ref{fig:results5000}. The three demonstrative frames are presented. One can notice that with the same design matrix the models demonstrate similar results. The figures show that $2000$ measurements can be used for object region detection, while $5000$ measurements which is only about $30\%$ of the input resolution are enough even to distinguish parts of the objects like doors and windows of the cars.

For the quantitative comparison of the results the following measures are used:
\begin{itemize}
\item \textit{Reconstruction error}: $ 
\dfrac{\|\mathbf{f}-\hat{\mathbf{f}}\|_{l_2}}{\|\mathbf{f}\|_{l_2}},$
where $\mathbf{f}$ is the signal ground truth, $\hat{\mathbf{f}}$ is the signal, reconstructed by the algorithm;
\item \textit{Background subtraction quality measure (BS quality)}: $\dfrac{|S( \mathbf{f} ) \cap S(\hat{\mathbf{f}})|}{|S( \mathbf{f}) \cup S(\hat{\mathbf{f}})|},$
where $S(\mathbf{f})$ is the ground truth foreground pixels, $S(\hat{\mathbf{f}})$ is the algorithm detected foreground pixels, $|\cdot|$ is the cardinality of the set;
\item \textit{Peak signal-to-noise ratio (PSNR)}: $10\log_{10}\left(\dfrac{\text{peakval}^2}{\text{MSE}}\right),$
where peakval is the maximum possible pixel value, that is 255 in our case. MSE is the mean square error between $\mathbf{f}$ and $\hat{\mathbf{f}}$;
\item \textit{Structural similarity index (SSIM)}: $
\dfrac{(2\mu{}_{\vphantom{\hat{\mathbf{f}}}\mathbf{f}}\mu{}_{\hat{\mathbf{f}}}+C_1)(2\sigma_{\mathbf{f}\hat{\mathbf{f}}}+C_2)}{(\mu_{\vphantom{\hat{\mathbf{f}}}\mathbf{f}}^2 + \mu_{\hat{\mathbf{f}}}^2+C_1)(\sigma_{\vphantom{\hat{\mathbf{f}}}\mathbf{f}}^2 + \sigma_{\hat{\mathbf{f}}}^2+C_2)},$
where $\mu_{\vphantom{\hat{\mathbf{f}}}\mathbf{f}}$, $\mu_{\hat{\mathbf{f}}}$, $\sigma_{\vphantom{\hat{\mathbf{f}}}\mathbf{f}}$, $\sigma_{\hat{\mathbf{f}}}$, $\sigma_{\mathbf{f}\hat{\mathbf{f}}}$ are the local means, standard deviations, and cross-covariance for the images $\mathbf{f}$, $\hat{\mathbf{f}}$ respectively, and $C_1, C_2$ are the regularisation constants
\end{itemize}
The difference between the uncompressed current frame $\mathbf{v}_k$ and the uncompressed background frame $\mathbf{b}$ is used as the ground truth signal $\mathbf{f}$ for every frame (the second columns in Figures \ref{fig:results2000} - \ref{fig:results5000}), since this is the signal which is compressed by (\ref{eq:linear_system}). 

\begin{figure*}[ht]
\centering
\subfloat[]{
\includegraphics{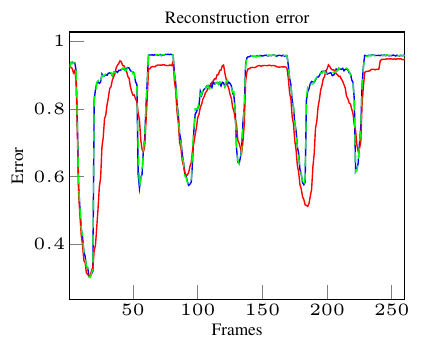}
}
\hfil
\subfloat[]{
\includegraphics{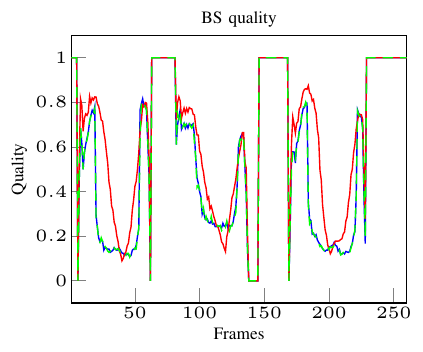}
}
\hfil
\subfloat[]{
\includegraphics{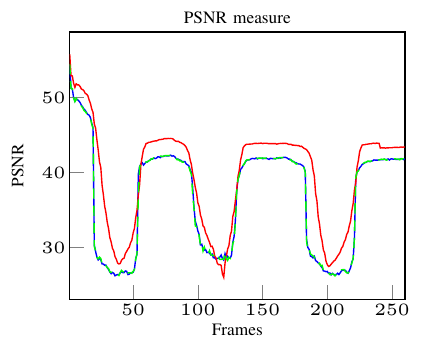}
}
\hfil
\subfloat[]{
\includegraphics{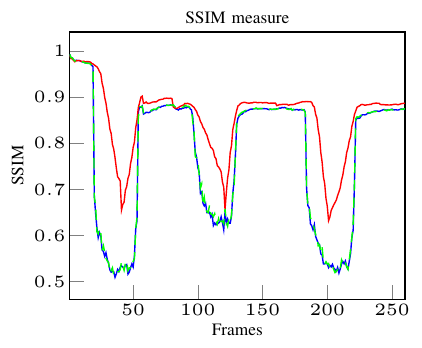}
}
\hfil
\subfloat[]{
\includegraphics{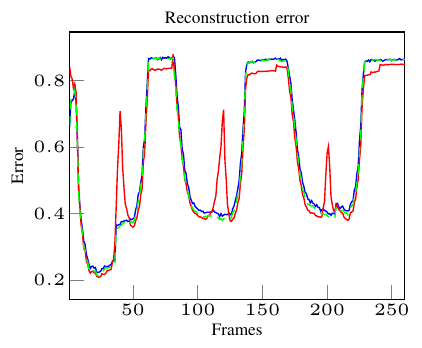}
}
\hfil
\subfloat[]{
\includegraphics{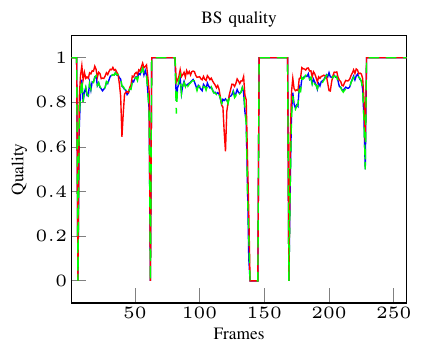}
}
\hfil
\subfloat[]{
\includegraphics{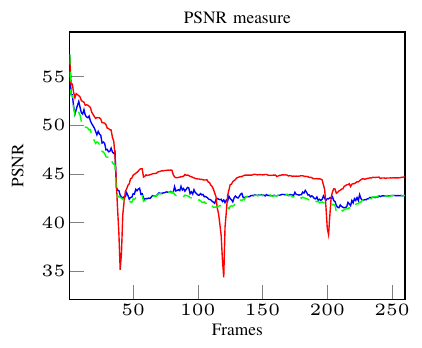}
}
\hfil
\subfloat[]{
\includegraphics{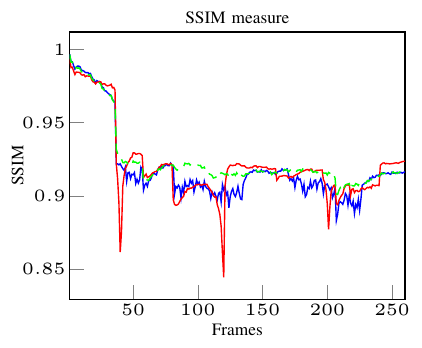}
}
\caption{Quantitative method comparison on the frame level. The blue line is for Bayesian compressive sensing, the red one is for multitask Bayesian compressive sensing, the dashed green one is for OMP. The top row corresponds to the set of the experiments with 2000 measurements, the bottom row corresponds to the set of the experiments with 5000 measurements. From left to right columns: the reconstruction error measure (values close to 1 refer to the frames without any foreground objects), the background subtraction quality measure, the PSNR measure, the SSIM measure.}
\label{fig:frame_comparison}
\end{figure*}

The results are presented in Figure~\ref{fig:frame_comparison}. All the quality measures -- reconstruction error, BS quality, PSNR and SSIM -- are calculated for every frame. The mean values among the frames for each measure can be found in Tables~\ref{tbl:2ktimes}~--~\ref{tbl:5ktimes}.

\begin{table}[!t]\footnotesize
  \caption{Method comparison based on 2000 measurements}
    \begin{tabular}{ | p{1cm} | p{1cm} | p{1cm} | p{1cm} | p{1cm} | p{1cm} | }
    \hline
    Algorithm & Mean frame reconstruction error & Mean frame BS quality  & Mean frame PSNR & Mean frame SSIM & Mean computational time (hours)\footnotemark\\ \hline
    BCS & 0.8037 & 0.3518 & 34.2007 & 0.7198  & 0.23\\ \hline
    Multitask BCS & 0.7608 & 0.4820 & 37.542 & 0.8384 & 0.67\\ \hline
    OMP & 0.8028 & 0.3510 & 34.1705 & 0.7204 & 0.51\\
    \hline
    \end{tabular}
\label{tbl:2ktimes}
\end{table}
 
\begin{table}[!t]\footnotesize
  \caption{Method comparison based on 5000 measurements }
    \begin{tabular}{ | p{1cm} | p{1cm} | p{1cm} | p{1cm} | p{1cm} |  p{1cm} | }
    \hline
    Algorithm & Mean frame reconstruction error & Mean frame BS quality & Mean frame PSNR & Mean frame SSIM & Mean computational time (hours)\footnotemark[\value{footnote}]\\ \hline
    BCS & 0.4713 & 0.8119 & 43.8251 & 0.9186 & 0.9\\ \hline
    Multitask BCS & 0.4702 & 0.8421 & 45.0028 & 0.9212 & 8.5\\ \hline
    OMP & 0.4578 & 0.8109 & 43.2720 & 0.9266 & 4.8\\
    \hline
    \end{tabular}
\label{tbl:5ktimes}
\end{table}
\footnotetext{The computational time is provided for a batch of 40 frames (BCS and OMP process each frame independently with 4 parallel workers, multitask BCS processes all 40 frames together). Implementation is made on the laptop  with i7-4702HQ CPU with 2.20GHz, 16 GB RAM using MATLAB 2015a}

Multitask Bayesian compressive sensing demonstrates the best results according to almost each measure. Bayesian compressive sensing and OMP show the competitive results but Bayesian compressive sensing works faster. It is worth to note that multitask Bayesian compressive sensing has the biggest variance among the runs with the different design matrices, while the variances of the Bayesian compressive sensing and OMP runs for the same matrices are quite small. 

\section{Conclusions and future work}
\label{sec:conclusion}
This work presents two Bayesian compressive sensing algorithms in the application of background subtraction. These are the applications of the conventional Bayesian compressive sensing and of the multitask Bayesian compressive sensing algorithms. The large size of the video frames leads to the high computational time for all methods, that is presented in Tables \ref{tbl:2ktimes} -- \ref{tbl:5ktimes}. However, the results presented in Figures~\ref{fig:results2000} -- \ref{fig:results5000} demonstrate the appropriate reconstruction quality of the original image based on only 5000 measurements (that is $\approx$~30\% of the original image size). 

The conventional Bayesian compressive sensing method demonstrates the similar results to the greedy algorithm OMP but BCS is more effective in terms of the computational time. If the computational time is not critical the extension of the Bayesian method designed for a multitask problem can improve the performance in terms of the different measures. Therefore other extensions of the Bayesian method to include the prior information need further research.   

The following problems can be addressed in future work.
Further research can be done on implementing different sparse Bayesian methods. The EP-based framework with the Laplace prior proposed in \cite{Seeger08} can be compared in terms of computational times and reconstruction errors. It uses the different inference scheme and prior, so the results should be different. Also the Markov Chain Monte Carlo (MCMC) \cite{Casella08} framework can be added to the comparison.

The current methods assume that the components of the foreground intensities are not correlated. For most cases the objects are grouped into several clusters, therefore more sophisticated sparsity models can be introduced to reflect the structure of the foreground. The Bayesian framework allows to implement such modifications. 

Exploring the applications in video tracking is one more avenue for further research.

\section{Acknowledgements}
The authors Olga Isupova and Lyudmila Mihaylova are grateful for the support provided by the EC Seventh Framework Programme [FP7 2013-2017] TRAcking in compleX sensor systems (TRAX) Grant agreement no.: 607400. Lyudmila Mihaylova acknowledges also the support from the UK Engineering and Physical Sciences Research Council (EPSRC) via the Bayesian Tracking and Reasoning over Time (BTaRoT) grant EP/K021516/1.

\bibliography{references/General,references/CompressiveSensing,references/NonparametricBayes,references/GraphicalModels}

\begin{thebibliography}{10}
\providecommand{\url}[1]{#1}
\csname url@samestyle\endcsname
\providecommand{\newblock}{\relax}
\providecommand{\bibinfo}[2]{#2}
\providecommand{\BIBentrySTDinterwordspacing}{\spaceskip=0pt\relax}
\providecommand{\BIBentryALTinterwordstretchfactor}{4}
\providecommand{\BIBentryALTinterwordspacing}{\spaceskip=\fontdimen2\font plus
\BIBentryALTinterwordstretchfactor\fontdimen3\font minus
  \fontdimen4\font\relax}
\providecommand{\BIBforeignlanguage}[2]{{%
\expandafter\ifx\csname l@#1\endcsname\relax
\typeout{** WARNING: IEEEtran.bst: No hyphenation pattern has been}%
\typeout{** loaded for the language `#1'. Using the pattern for}%
\typeout{** the default language instead.}%
\else
\language=\csname l@#1\endcsname
\fi
#2}}
\providecommand{\BIBdecl}{\relax}
\BIBdecl

\bibitem{Bach2012}
F.~Bach, R.~Jenatton, J.~Mairal, and G.~Obozinski, ``Optimization with
  sparsity-inducing penalties,'' \emph{Found. Trends Mach. Learn.}, vol.~4,
  no.~1, pp. 1--106, Jan. 2012.

\bibitem{Candes2008intro}
E.~Candes and M.~Wakin, ``An introduction to compressive sampling,''
  \emph{Signal Processing Magazine, IEEE}, vol.~25, no.~2, pp. 21--30, March
  2008.

\bibitem{Mihaylova2014}
A.~Y. Carmi, L.~S. Mihaylova, and S.~J. Godsill,
  ``\BIBforeignlanguage{English}{Introduction to compressed sensing and sparse
  filtering},'' in \emph{\BIBforeignlanguage{English}{Compressed Sensing and
  Sparse Filtering}}, ser. Signals and Communication Technology, A.~Y. Carmi,
  L.~Mihaylova, and S.~J. Godsill, Eds.\hskip 1em plus 0.5em minus 0.4em\relax
  Springer Berlin Heidelberg, 2014, pp. 1--23.

\bibitem{Tibshirani96}
R.~Tibshirani, ``\BIBforeignlanguage{English}{Regression shrinkage and
  selection via the lasso},'' \emph{\BIBforeignlanguage{English}{Journal of the
  Royal Statistical Society. Series B (Methodological)}}, vol.~58, no.~1, pp.
  267--288, 1996.

\bibitem{Seeger08}
M.~Seeger, ``Bayesian {I}nference and {O}ptimal {D}esign in the {S}parse
  {L}inear {M}odel,'' \emph{Journal of {M}achine {L}earning {R}esearch},
  vol.~9, pp. 759--813, 2008.

\bibitem{Tipping2001}
M.~E. Tipping, ``Sparse bayesian learning and the relevance vector machine,''
  \emph{The journal of machine learning research}, vol.~1, pp. 211--244, 2001.

\bibitem{Carin2008bcs}
S.~Ji, Y.~Xue, and L.~Carin, ``Bayesian compressive sensing,'' \emph{IEEE
  Transactions on Signal Processing}, vol.~56, no.~6, pp. 2346--2356, June
  2008.

\bibitem{Seeger08cs}
M.~W. Seeger and H.~Nickisch, ``Compressed sensing and bayesian experimental
  design,'' in \emph{Proceedings of the 25th International Conference on
  Machine Learning}, ser. ICML '08.\hskip 1em plus 0.5em minus 0.4em\relax New
  York, NY, USA: ACM, 2008, pp. 912--919.

\bibitem{Bach14}
J.~Mairal, F.~R. Bach, and J.~Ponce, ``Sparse modeling for image and vision
  processing,'' \emph{CoRR}, vol. abs/1411.3230, 2014.

\bibitem{Cevher2008}
V.~Cevher, A.~Sankaranarayanan, M.~F. Duarte, D.~Reddy, and R.~G. Baraniuk,
  ``Compressive sensing for background subtraction,'' in \emph{European Conf.
  Comp. Vision (ECCV)}, 2008, pp. 155--168.

\bibitem{Warnell2014}
G.~{Warnell}, S.~{Bhattacharya}, R.~{Chellappa}, and T.~{Basar},
  ``{Adaptive-Rate Compressive Sensing Using Side Information},'' \emph{ArXiv
  e-prints}, Jan. 2014.

\bibitem{Baraniuk08}
R.~Baraniuk, M.~Davenport, R.~DeVore, and M.~Wakin,
  ``\BIBforeignlanguage{English}{A simple proof of the restricted isometry
  property for random matrices},''
  \emph{\BIBforeignlanguage{English}{Constructive Approximation}}, vol.~28,
  no.~3, pp. 253--263, 2008.

\bibitem{Murphy2012}
K.~P. Murphy, \emph{Machine Learning: A Probabilistic Perspective}.\hskip 1em
  plus 0.5em minus 0.4em\relax The MIT Press, 2012.

\bibitem{Carin2009mcs}
S.~Ji, D.~Dunson, and L.~Carin, ``Multitask compressive sensing,'' \emph{IEEE
  Transactions on Signal Processing}, vol.~57, no.~1, pp. 92--106, Jan 2009.

\bibitem{Mallat93}
S.~Mallat and Z.~Zhang, ``Matching pursuits with time-frequency dictionaries,''
  \emph{IEEE Transactions on Signal Processing}, vol.~41, no.~12, pp.
  3397--3415, Dec 1993.

\bibitem{Casella08}
T.~Park and G.~Casella, ``The bayesian lasso,'' \emph{Journal of the American
  Statistical Association}, vol. 103, no. 482, pp. 681--686, 2008.

\end{thebibliography}

\end{document}